# Imitation Learning of Compression Pattern in Robotic-Assisted Ultrasound Examination Using Kernelized Movement Primitives

Diego Dall'Alba, Lorenzo Busellato, Thiusius Rajeeth Savarimuthu, Zhuoqi Cheng, *Member, IEEE*, and Iñigo Iturrate, *Member, IEEE*

*Abstract*—Vascular diseases are commonly diagnosed using Ultrasound (US) imaging, which can be inconsistent due to its high dependence on the operator's skill. Among these, Deep Vein Thrombosis (DVT) is a common yet potentially fatal condition, often leading to critical complications like pulmonary embolism. Robotic US Systems (RUSs) aim to improve diagnostic test consistency but face challenges with the complex scanning pattern requiring precise control over US probe pressure, such as the one needed for indirectly detecting occlusions during DVT assessment. This work introduces an imitation learning method based on Kernelized Movement Primitives (KMP) to standardize the contact force profile during US exams by training a robotic controller using sonographer demonstrations. A new recording device design enhances demonstration acquisition, integrating with US probes and enabling seamless force and position data recording. KMPs are used to link scan trajectory and interaction force, enabling generalization beyond the demonstrations. Our approach, evaluated on synthetic models and volunteers, shows that the KMP-based RUS can replicate an expert's force control and US image quality, even under conditions requiring compression during scanning. It outperforms previous methods using manually defined force profiles, improving exam standardization and reducing reliance on specialized sonographers.

Received 16 May 2024; revised 11 July 2024 and 25 August 2024; accepted 22 September 2024. Date of publication 3 October 2024; date of current version 12 November 2024. This article was recommended for publication by Associate Editor T. Haidegger and Editor P. Dario upon evaluation of the reviewers' comments. This work was supported in part by the Odense Robotics Control and Learning of Contact Transitions (CoLeCT) Project, funded by the Danish Ministry of Higher Education and Science, and in part by the SDU I4.0 Lab at the University of Southern Denmark. *(Corresponding author: Diego Dall'Alba.)*

This work involved human subjects or animals in its research. The authors confirm that all human/animal subject research procedures and protocols are exempt from review board approval.

Diego Dall'Alba is with the Altair Robotics Lab, Department of Computer Science, University of Verona, 37123 Verona, Italy, and also with the SDU Robotics, The Maersk Mc-Kinney Moller Institute, University of Southern Denmark, 5230 Odense, Denmark (e-mail: diego.dallalba@univr.it).

Lorenzo Busellato is with the Altair Robotics Lab, Department of Computer Science, University of Verona, 37123 Verona, Italy.

Thiusius Rajeeth Savarimuthu is with the SDU Robotics, the Maersk Mc-Kinney Moller Institute, University of Southern Denmark, 5230 Odense, Denmark, and also with Ropca ApS, 5260 Odense, Denmark.

Zhuoqi Cheng and Iñigo Iturrate are with the SDU Robotics, The Maersk Mc-Kinney Moller Institute, University of Southern Denmark, 5230 Odense, Denmark.

## I. INTRODUCTION

CARDIOVASCULAR diseases are prevalent, and prompt diagnosis and treatment are crucial for managing these conditions effectively and preventing life-threatening complications [1]. Ultrasound (US) imaging is a widely used diagnostic tool since it provides real-time imaging without ionizing radiation and can be performed with portable and cost-effective equipment [2], [3]. However, vascular scanning demands advanced visuo-tactile skills, typically mastered by experienced sonographers through extensive training [2]. For accurate anatomical and pathological assessments, such as evaluating atherosclerosis severity, it is vital to perform the US scan without deforming the blood vessel [4].

Another pathology that requires complex US scan patterns for diagnosis is Deep Vein Thrombosis (DVT), a common condition of the extremities characterized by blood clots that block venous return, which can lead to serious issues like pulmonary embolism [5]. Although US images do not typically visualize the thrombus directly, they can reveal its presence through changes in vein compressibility. During the scan, the sonographer applies pressure with the US probe at regular intervals along the entire course of the examined vein. Healthy veins will compress under this pressure, indicating no obstruction. Conversely, a vein that resists compression suggests the presence of a thrombus, signaling a potential DVT. The effectiveness of this diagnostic method heavily relies on the practitioner's expertise, affecting its consistency and standardization. Since DVT is a chronic condition, patients at risk or those already diagnosed should undergo recurrent scans to monitor the onset or progression of the disease. However, implementing such screening or follow-up programs is challenging due to the burden on medical personnel [3].

To address these challenges, Robotic Ultrasound Systems (RUSs) have emerged as a solution that can standardize and automate US scans [2], [6]. One of the simplest ways to control RUSs is through a teleoperation scheme, where the operator remotely controls a robotic manipulator with an US probe as an end-effector [2]. This scheme enables US imaging on patients in remote locations or in situations requiring interpersonal distancing, such as during the COVID-19 pandemic [7]. While it has the potential to reduce the physical workload for the operator, an optimized multimodal interaction interface needs to be developed to ensure its effective use [8]. This requires training in the use of the RUS in addition to the standard US imaging training.

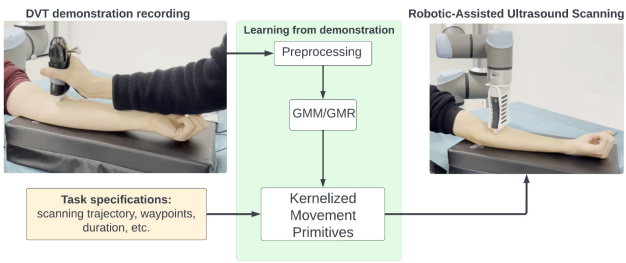

Fig. 1. High-level diagram showing the proposed approach based on Kernelized Movement Primitives to learn from sonographer demonstrations the correct scanning patterns for a Robotic US system.

Therefore, an interesting research area is to increase the autonomy of RUS and decrease the reliance on expert sonographers [9], [10]. As learning the dependence between scan location, pressing force, and quality of the US image is a non-trivial task, many approaches have adopted machine learning methods. Broadly, these tend to fall between reinforcement learning (RL), imitation learning (IL), and optimization-based methods, such as Bayesian Optimization (BO). A more detailed overview and comparison of these methods will be provided in Section II.

In this work, we focus on IL, as this family of methods focuses on directly transferring skills from demonstrations provided by an expert in the task to the robot. Given the complexity of the US vascular scan, IL approach can be well-suited since we can then formulate the problem as learning scanning skills directly from sonographers' demonstrations.

One of the critical aspect of an IL pipeline is capturing experts' demonstrations. A popular method is kinesthetic teaching [11], [12], [13], where the user moves the robotic arm to provide demonstrations. However, the ergonomics differ significantly from those of the probe, causing the expert's manual skills to transfer imprecisely, resulting in suboptimal demonstrations and learning [14].

While existing IL approaches to US scanning have considered learning trajectories [15] or optimizing them based on US images [13], not much emphasis has been put on learning the scanning force profile. This is crucial in conditions like DVT, where learning how to adapt the force profile of the scan to the characteristics of each patient is fundamental for optimal scanning, as different thicknesses of limb and locations of vessels will require different scanning forces.

In this work, we address most of these limitations associated with prior IL methods for US scanning involving complex interaction force patterns. The innovative contributions include:

1) We introduce a demonstration acquisition setup that allows the recording of US images, interaction forces and torques together with position and orientation of the US probe without significantly altering the probe's ergonomics (thus, making it more natural to demonstrate for expert sonographers). We publicly share the hardware designs and supporting software to facilitate the adoption of the proposed approach (https://github.com/lbusellato/CoLeCT).

2) We adopt Kernelized Movement Primitives [16] for imitation learning of vascular RUS. Similar to other IL methods that have previously been used in RUS, such as Gaussian Mixture Models (GMM) and Gaussian Mixture Regression (GMR) [15], [17], [18] or Interaction probabilistic movement primitives (iProMP) [19], KMPs encode the probability distribution over robot trajectories covered by the demonstrations. Unlike these methods, KMPs kernelize this representation, allowing them to condition the output on high-dimensional inputs. This is relevant for RUS demonstrations, where positions, orientations, forces and torques features all play a role in a successful task execution (i.e., obtaining suitable US image quality). In this work, we condition the output force of the KMP depending on the relative scanning position to allow adaptation to patient-specific vascular anatomy.

3) We extensively validate our approach using realistic anatomical models and healthy volunteers to evaluate the ability of KMPs to replicate both the forces and the quality of images obtained by the human operator.

While our proposed framework in its current form is not fully capable of scanning for DVT, we are heavily inspired by the requirements imposed by this disease on the scanning force pattern, i.e., the need to adapt the force profile to achieve proper vessel compression. As previously mentioned, DVT requires a patient to undergo multiple scans to assess disease progression. We therefore envision the medical professional performing the first scan. This scan can then recorded and used to train the system such that it is able to autonomously perform the follow-up scans for the same patient. Here, our system's ability to automatically learn and adapt the force profile is beneficial, as it prevents the medical professional from having to manually define target force values, which they might not have the technical expertise to perform.

The rest of this article is organized as follows: Section II presents the relevant works in the state of the art; Section III provides background information to understand the method described in Section IV; Section V details the experimental setup; Sections VI and VII contain the description and discussion of the results, respectively; and Section VIII provides the conclusions.

## II. RELATED WORKS

This section reviews the relevant literature, particularly, the use of robot in medical US applications. As mentioned previously, to bypass the complexity of modeling the US examination procedure directly, many RUS approaches have turned to learning or optimization. Here, we will limit our review to these approaches. For a more comprehensive survey of the RUS field, the reader can consult the recent review papers [2], [9], [20].

## A. Reinforcement Learning

The main idea behind RL is to define a *reward function* that assigns high scores to optimal actions and low scores to sub-optimal ones. The robot then explores actions on its own and is encouraged to learn actions that maximize the reward. In US scanning task, RL involves exploring sequences of probe movements based on the current observations (e.g., US images, force, probe pose) to plan the next step. Some works have used image-based RL [21], [22] to guide the US probe for scanning spine anatomy, using image features as inputs. However, they did not account for the skin-probe contact during the scanning. Another work [23] used a deep-learning approach combining US image features with interaction forces and multi-modal data from experts. They used a guided exploration technique to improve the model's generalization, where the user can correct the model's actions. A similar deep learning approach [24] replaced the guided exploration with a sampling-based method that considered the predicted image quality during the US scanning. Ning et al. used force-based RL [25], [26] to control the probe torques around two axes, while an admittance controller was integrated for the other degrees of freedom. In [26], the same authors proposed a deep learning approach to control the probe position based on image features, while force-based RL was used for the probe orientation, as in the previous work.

## B. Imitation Learning

The main limitation of RL approaches is the need to define reward functions and perform many trials. These trials often need to be performed in a simulated or simplified setting to avoid unsafe behaviors, making the method hard to transfer to realistic conditions and hindering its ability to generalize. Thus, an alternative strategy is to approach the learning in a supervised manner, either attempting to learn the expert's reward function (Inverse Reinforcement Learning, IRL) and subsequently applying RL, or directly attempting to map observations of the environment to actions based on expert demonstrations. Here, we survey both families of methods.

*1) Inverse Reinforcement Learning:* A challenge of the IRL approach is the quality of the demonstrations, which often contain exploratory and sub-optimal actions in RUSs. To solve this problem, probabilistic ranking mechanisms have been suggested, based on temporal [12] or spatial [13] information. In [12], the authors gave more weights to the data collected at the end of the demonstrations, which were less exploratory than the initial ones, and introduced a temporal ranking mechanism. In [13], the authors used a similarity measure between images based on mutual information and the assumption that the areas of interest were visited several times during the scan. With the probabilistic ranking mechanism, a training dataset can be created from a few demonstrations and used to train a reward inference mechanism. The inferred reward was used in a BO or RL approach for automatic probe navigation. Another IRL approach was proposed in [27], as an extension of [25]. In this work, demonstrations were created with an RL approach with expert user feedback to ensure optimality, and then IRL was used to learn the reward with a maximum entropy approach.

*2) Movement Primitives:* Alternatively, a direct mapping between observations and actions can be learned. This is the case with behavioral cloning and movement primitives. A seminal work was [15], which used GMM and GMR to learn RUS from demonstrations. This work only used trajectories of positions and orientations, and ignored interaction forces and US images. Deng et al. [18] used GMM/GMR with probe orientation, interaction forces, and a feature vector from US images to learn from expert sonographer demonstrations. This work did not use the translation component of the probe, since the probe pose during the demonstrations was measured with an IMU sensor. A work that learned from multiple trajectories was [28], which used GMM/GMR to fit the user demonstrations. The fitting result was given to Dynamical Movement Primitives (DMPs) to generate the trajectories during replication. This work focused on the force control without acquiring US images, limiting learning to constant forces normal to the surface and restricted translational components, while maintaining a fixed probe orientation. In [19], iProMP was used with a fuzzy approach to adapt the generated trajectories. The model used the trajectories, quality index, and features of US images. The learned model, based on the image information, could update the robot's position during the reproduction phase.

## C. Bayesian Optimization (BO)

A widely used approach in RUS is based on BO, a method that efficiently searches for the best parameters of a closed box function. Goel et al. first proposed BO for a two-dimensional planning problem in RUS, where the goal was to scan the area with the highest concentration of vascular structures [29]. They used a reward based on segmenting vascular structures, and a hybrid force controller to maintain a constant interaction force and a normal probe orientation. BO was also used to learn adjustments to the orientation of the US probe based on force-sensor data to position it normal to the surface being scanned to achieve maximum image quality [30]. Raina et al. extended the previous works to a three-dimensional (3D) planning problem, where they learned the normal force profile along with the other two directions [31]. They also incorporated expert demonstrations (obtained with a teleoperation approach) as priors in the BO process. Additionally, the same research group improved the reward mechanism, using an image quality estimate based on convolutional neural networks from [32]. The use of quality indices or feature vectors is essential for BO approaches, since they cannot handle directly high-dimensional inputs such as US images.

## D. Our Approach

In contrast to previously described works, this work focuses on learning the variable force profiles of a RUS, such as the ones required for conducting an effective US DVT examination. This type of scan requires learning more complex force profiles than those considered in previous works for other vascular districts (e.g., US examination of the carotid artery [4], [26]). Indeed, in most previous RUS researches, the interaction force is considered constant [4], [13], [20], [29], [31], [32].

Our choice of movement primitive representation is motivated by the desire for certain properties: 1) a probabilistic representation in order to capture variations in US scanning trajectories, where multiple valid approaches for the scanning motion can be chosen. This excludes Dynamic Movement Primitives as a choice. 2) Start-, end-, and via-point modulation so as to simplify the generalization of learned trajectories to specific conditions for each patient. This excludes GMM/GMR as a choice. 3) The possibility to condition the model output on high-dimensional inputs. This is in order to allow for future extension of the framework to handle direct input of US images. This excludes Probabilistic Dynamic Movement Primitives (ProDMP/iProDMP) as a choice.

For the above reasons, KMP is chosen, as they present all of the required properties. Gaussian Process Regression [33] would also exhibit similar properties, but is a lesser-known approach within LfD. Other approaches, such as ProDMP [34], while presenting some other advantages, require large amounts of data to train, which could present problems in our clinical scenario.

DVT US scanning has not been extensively researched due to the procedural complexity for clinicians and, consequently, for RUS. Pioneering work focused on the combined analysis of image features and interaction forces but was limited to manual scanning [35]. Subsequent research shifted towards high-level planning using RGB-D data, but it did not address the learning of force profiles necessary for compression maneuvers [36]. Recently, Huang et al. introduced a method that merges RGB-D data with imaging features to determine an optimal high-level scanning trajectory [5]. This trajectory was then used to generate a virtual fixture that streamlined the user interaction and standardized the examination process. While this approach offered a standardized DVT scanning protocol, it lacked a learning-from-demonstration component, thus still necessitating a human operator. Instead, our work focuses on learning from demonstrations exploiting KMP to automate the scanning process and alleviate the burden on operators.

## III. Background

This section presents the required background on GMM and GMR, which are used to model the input demonstrations probabilistically as a pre-processing step, and KMPs, which are used to encode the demonstrations for reproduction.

### A. Probabilistic Modeling of the Demonstrations

During the first phase, human expert demonstrations are captured for the target US scanning procedure. These demonstrations include the Cartesian position and orientation of the US probe, and the associated contact forces and torques. These demonstrations are first pre-processed in order to address the different sampling rates of the various sensors used in the system (e.g., force-torque and pose tracking) by interpolating missing data points. In addition, subsampling may be applied to reduce the dataset complexity. Since different demonstrations naturally express the same skill under different time- and velocity-profiles, we use Soft-Dynamic Time Warping (Soft-DTW) to temporally align the different recordings [37]. The processed demonstrations are then decomposed, extracting the task-relevant input and output features (e.g., relative positions and interaction forces), and finally collected in a demonstration database:

$$\{\{s_{n,h}, \xi_{n,h}\}_{n=1}^{N}\}_{h=1}^{H},$$

where $s_{n,h}$ and $\xi_{n,h}$ are, respectively, the $n$-th $\mathcal{I}$-dimensional input and $\mathcal{O}$-dimensional output vectors of the $h$-th demonstration in the database. N and H denote the length of each demonstration and the number of demonstrations in the database, respectively.

As mentioned in Section II-D, the KMP representation was chosen for its probabilistic properties and ability to modulate the output trajectory by adding via-points. KMP requires a probabilistic representation of the demonstrations as an input. Therefore, as suggested in [16], we approximate the demonstration distribution as a Gaussian mixture model (GMM):

$$p(s|\xi) \approx \sum_{c=1}^{C} \pi_c \mathcal{N}(s|\mu_c, \Sigma_c), \quad (1)$$

where $C$ is the number of Gaussian components, $\pi_c$ are the prior probabilities, and $\mathcal{N}(s|\mu_c, \Sigma_c)$ are the Gaussian distributions with mean $\mu_c$ and covariance $\Sigma_c$. The parameters of the GMM are learned using the expectation-maximization (EM) algorithm. The joint probability distribution encoded by the GMM is decomposed as:

$$\mu_c = \begin{bmatrix} \mu_c^s \\ \mu_c^\xi \end{bmatrix} \quad \Sigma_c = \begin{bmatrix} \Sigma_c^{ss} & \Sigma_c^{s\xi} \\ \Sigma_c^{\xi s} & \Sigma_c^{\xi\xi} \end{bmatrix}. \quad (2)$$

Then, the conditional expectation for a new input $\hat{s}$ can be computed via GMR. The conditional expectation for the mean associated with the new input is computed as follows:

$$\hat{\mu}_n = \mathbb{E}\left[\hat{\xi}_n \mid \hat{s}\right] = \sum_{c=1}^{C} h_c(\hat{s}) \, \mu_c(\hat{s}), \quad (3)$$

where $h_c(\hat{s})$ is the posterior probability of the $c$-th Gaussian component given the input, which is defined as:

$$h_c(\hat{s}) = \frac{\pi_c \mathcal{N}(\hat{s}|\mu_c, \Sigma_c)}{\sum_{j=1}^{C} \pi_j \mathcal{N}(\hat{s}|\mu_j, \Sigma_j)}, \quad (4)$$

where $\mu_c(s_n) = \mu_c^\xi + \Sigma_c^{\xi s}(\Sigma_c^{ss})^{-1}(\hat{s} - \mu_c^s)$. In the same way, the conditional expectation of the covariance is computed as:

$$\hat{\Sigma}_n = \mathbb{D}\left[\hat{\xi}_n \mid \hat{s}\right] = \mathbb{E}\left[\hat{\xi}_n \hat{\xi}_n^T\right] - \mathbb{E}\left[\hat{\xi}_n \mid \hat{s}\right] E\left[\hat{\xi}_n \mid \hat{s}\right]^T, \quad (5)$$

where

$$\mathbb{E}\left[\hat{\xi}_n \xi_n^T\right] = \sum_{c=1}^{C} h_c(s_n)\left(\bar{\Sigma}_c + \mu_c(s_n)\mu_c(s_n)^T\right) \quad (6)$$

and $\bar{\Sigma}_c = \Sigma_c^{\xi\xi} - \Sigma_c^{\xi s}(\Sigma_c^{ss})^{-1}\Sigma_c^{s\xi}$.

*B. Learning From Demonstration*

In the learning phase, a reference database is generated through GMR applied on a reference input trajectory. The learning is carried out by exploiting KMP to define a parametric trajectory that can be approximated as a Gaussian distribution of unknown mean $\mu_w$ and covariance $\Sigma_w$:

$$\xi(s) = \Theta(s)^T w \sim \mathcal{N}\Big(\Theta(s)^T \mu_w, \Theta(s)^T \Sigma_w \Theta(s)\Big), \quad (7)$$

where $\Theta(s) = \varphi(s) I_\mathcal{O}$ is a matrix of $\mathcal{B}$-dimensional basis functions $\varphi(s) \in \mathbb{R}^\mathcal{B}$ and $w \in \mathbb{R}^{\mathcal{BO}}$ is a weight vector. To derive KMP's formulation, the following objective function is defined:

$$J(\mu_w, \Sigma_w) = \sum_{n=1}^{\mathcal{N}} D_{KL}\big(\mathcal{P}_p(\xi \mid s_n) \mid\mid \mathcal{P}_r(\xi \mid s_n)\big), \quad (8)$$

where $\mathcal{P}_p(\xi \mid s_n)$ and $\mathcal{P}_r(\xi \mid s_n)$ are the probabilistic distributions of the parametric trajectory and of the reference trajectory respectively, given the input $s_n$. The Kullback-Leibler divergence, denoted with $D_{KL}(\cdot \mid\mid \cdot)$, is a measure of information loss between distributions. As such, deriving KMP requires minimizing $J(\mu_w, \Sigma_w)$ with respect to the unknown $\mu_w$ and $\Sigma_w$. By leveraging the characteristics of KL-divergence between Gaussian distributions and the parallels in minimization sub-problems seen in Kernel Ridge Regression, we can deduce expressions for the unknown mean and covariance. The key observation from these findings is that basis functions exist only in the form of inner products among them. This observation paves the way for KMP to incorporate kernelization, circumventing the complicated and resource-intensive task of explicitly defining the basis functions, especially when considering high-dimensional inputs.

The selection of the kernel function, denoted as $k(\cdot, \cdot)$, is fundamental. A guiding principle for this choice is its congruence with the statistical properties of the underlying data. Given that KMP's formulation assumes data to follow normal distributions, we have opted for the Radial Basis Function (RBF) kernel, defined as:

$$k(s_i, s_j) = e^{-\sigma_f \|s_i - s_j\|^2}, \quad (9)$$

where $\sigma_f$ is a regularization factor. The RBF kernel constructs the kernel matrix $K$ as follows:

$$K = \begin{bmatrix} k(s_1, s_1) & k(s_1, s_2) & \ldots & k(s_1, s_N) \\ k(s_2, s_1) & k(s_2, s_2) & \ldots & k(s_2, s_N) \\ \vdots & \vdots & \ddots & \vdots \\ k(s_N, s_1) & k(s_N, s_2) & \ldots & k(s_N, s_N) \end{bmatrix}. \quad (10)$$

Similarly, given a new input $s^*$, the kernel matrix $k^*$ is constructed as follows:

$$k^* = \begin{bmatrix} k(s^*, s_1) & k(s^*, s_2) & \ldots & k(s^*, s_N) \end{bmatrix}. \quad (11)$$

The kernelized mean expectation of KMP for the new input is then

$$\mu_w(s^*) = k^* (K + \lambda \Sigma)^{-1} \mu. \quad (12)$$

The covariance expectation for the input $s^*$ can be similarly kernelized as

$$\Sigma_w(s^*) = \frac{N}{\lambda_c}\Big(k(s^*, s^*) - k^*(K + \lambda \Sigma)^{-1} k^{*T}\Big), \quad (13)$$

where $\lambda$ and $\lambda_c$ are regularization factors introduced to avoid overfitting the mean and covariance predictions, respectively.

$$\Sigma = \text{blockdiag}\Big(\hat{\Sigma}_1, \hat{\Sigma}_2, \ldots, \hat{\Sigma}_N\Big)$$
$$\mu = \begin{bmatrix} \hat{\mu}_1^T & \hat{\mu}_2^T & \ldots & \hat{\mu}_N^T \end{bmatrix}^T. \quad (14)$$

*C. Demonstration Reproduction*

In the reproduction phase, the model is queried with the same reference trajectory used during training. However, it is possible to introduce constraints, such as new waypoints or different start/end points. Given a desired point $\{s_d, \mu_d, \Sigma_d\}$, the model compares its input $s_d$ against all inputs in the reference database to find the nearest point $s_r$, i.e., the one that satisfies $d(s_d, s_r) \leq d(s_d, s_n) \forall n \neq r$ with $d(\cdot)$ being the Euclidean distance function. If such a point is found, it is replaced by the desired point if the distance between them is smaller than a user-defined threshold. Otherwise, the desired point is appended to the reference database.

In our experiments, $r = 5 \cdot 10^{-4}$, which resulted in always successfully finding a correspondence and no new points added to the database. In practice, the choice of $r$ can be determined experimentally, and depends on both the dataset dimensionality $N$ and the accuracy requirements, as replacing points in the database will naturally lead to the original demonstration points being disregarded in the probabilistic representation.

## IV. METHODS

The proposed RUS framework is structured in three phases: probabilistic modeling of the demonstrations, learning from demonstration and reproduction.

Figure 2 shows a block diagram of the RUS system that we propose in this work. The system consists of three hierarchical processes. The highest-level process deals with using the task specifications to generate a reference trajectory for the scan, based on manual input [38], [39], three-dimensional reconstructions of the patient's body surface [5], [36] or pre-operative data [40], [41]. The process also extracts any user-defined trajectory waypoints, as well as generating a reference trajectory for the training of KMP, which is generally the same as the one for the actual scan. The middle-level process deals with constructing the reference database from the recorded demonstrations, its probabilistic encoding using GMM/GMR and the training of the KMP model, i.e., the computation of the kernel matrices. The bottom-level process, based on the trained KMP model, controls the probe during the US scan, following the trajectory from the upper-level processes. It has to account for the interaction between the probe and the patient's body, a complex task involving tissue deformation and force control. We focus on this lower-level process in this work and assume that the higher level process can be implemented by any existing methods [5], [36],

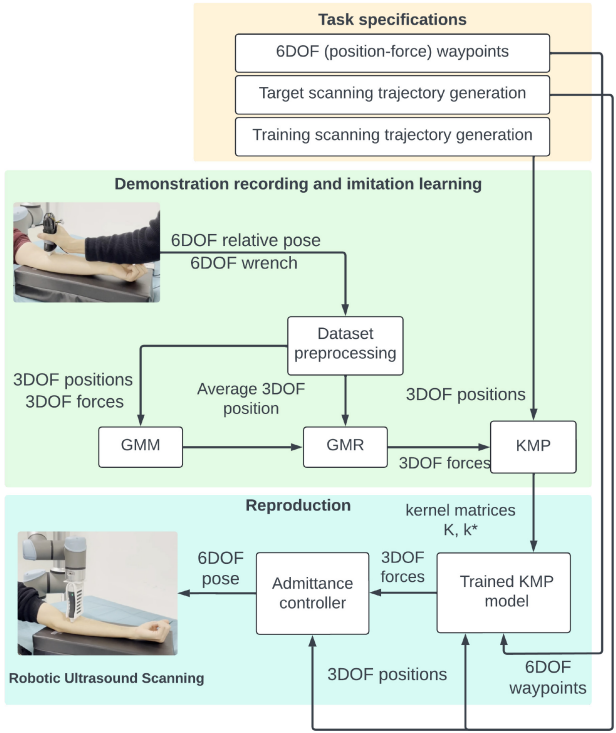

Fig. 2. Reference model of the RUS system, describing the three main processes involved in learning from demonstration.

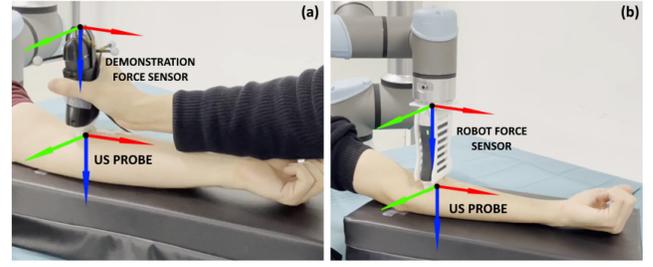

Fig. 3. Detail of the reference systems involved in the acquisition setup (a) and during reproduction (b). The additional reference systems of the tracking system (left) and the robot base (right) are not shown in the figure.

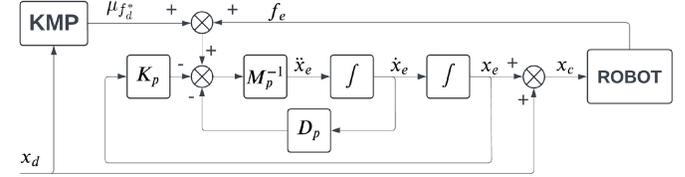

Fig. 4. Block diagram of the admittance controller used in this work.

[40], [41]. Unless stated otherwise, we manually select the upper-level plan (i.e., target pose or desired scan trajectory) throughout the work.

### A. Force Reference Generation

To generate the reference forces for task reproduction, we query the KMP at each timestep based on the current state of the robot, i.e.:

$$p(f_d^* | x_d^*) \sim \mathcal{N}\left(\mu_{f_d^*}(x_d^*), \Sigma_{f_d^*}(x_d^*)\right), \quad (15)$$

where $\mu_{f_d^*}(x_d^*)$, $\Sigma_{f_d^*}(x_d^*)$ are computed using Eqs. 12 and 13, $x_d^*$ is the desired pose of the robot end-effector used to query the KMP, and $f_d^*$ is the reference (desired) force distribution outputted by the KMP.

### B. Robot Control

To control the robot, the mean force for the given state, $\mu_{f_d^*}$, queried from the KMP as described in the previous section, is sent as a reference to a Cartesian admittance controller [42] with an added feedforward force reference:

$$M_p \ddot{x}_e + D_p \dot{x}_e + K_p x_e = h_e + \mu_{f_d^*}, \quad (16)$$

where $M_p, D_p, K_p \in \mathbb{R}^{3\times 3}$ are positive definite (semidefinite in the case of $K_p$) matrices describing the virtual mass, damping and stiffness of the robot end-effector, respectively. $h_e \in \mathbb{R}^3$ is the measured force at the end-effector, $\mu_{f_d^*} \in \mathbb{R}^3$ is a feedforward target force, corresponding to the mean of the output queried from the KMP, and $x_e \in \mathbb{R}^3$ and defined as $x_e = x_c - x_d$ is the error between the compliant and desired frames, where the compliant frame, $x_c$, corresponds to the position of the robot end-effector. The orientation is described in a manner analogous to Eq. (16) using quaternions.

To achieve a hybrid control scheme that ensures tracking of the target force in the axis of contact of the probe and tracking of position in the other two axes, we further define

$$K_p = \text{diag}(K_x, K_y, 0), \quad (17)$$

where $K_x$ and $K_y$ are the desired stiffness for position tracking in the $x$- and $y$-axes, respectively, of the end-effector frame. The frames of reference are shown in Figure 3.

A control diagram of the system, including generation of the KMP reference trajectories for the position and force, can be seen in Figure 4.

### C. Demonstration Recording Setup

The recording of US scanning demonstrations is a fundamental step in IL approaches, as the RUS capabilities rely on these demonstrations. Previous methods are based on kinesthetic teaching [12], [13] or teleoperation [31], but this limits transferring the sonographer's visuotactile skills due to significant differences in control interfaces compared to clinical practice. Other works propose the integration of force sensors, position tracking sensors and US probe in a non-robotic compact setup [15], [24], enhancing the sonographer's intuitive control during demonstrations. However, these solutions compromise ergonomics for sensor integration and accurate data recording. Our approach combines design optimization without sacrificing demonstration recording performance. Specifically, the designed handheld recorder allows the sonographer to perform US scanning close to the clinical practice.

The US probe was a Clarius HD3 Linear probe, which was connected to the acquisition workstation via Wi-Fi. In order to record the applied force, a force sensor (WITTENSTEIN/WIKA Resense HEX 12) was used. This force sensor is very light and compact, but provides a measurement range of 25N with 1% accuracy. The sensor connected two 3D printed adapters, showed in Figure 5, namely the cap and the sleeve. The cap, attached on top of the probe, housed the optical markers used for pose tracking. The sleeve was designed to replicate the profile of the probe's body, ensuring ergonomic consistency. It was loosely fitted around the probe to allow the accurate transfer of user-applied forces to the force sensor. Thanks to the Optitrack PrimeX system with Motive software suite, the 6-DoF pose of the US probe could be tracked via the optical markers in real time.

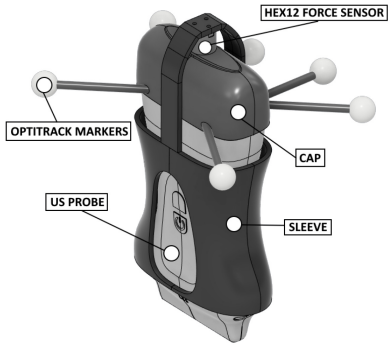

Fig. 5. Rendering of the US probe with the 3D printed adapter that integrates the force sensor and the rigid body to house the optical markers of the pose tracking system.

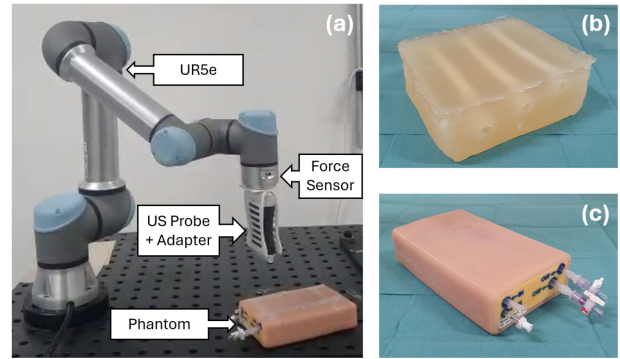

Fig. 6. (a) Main components of the RUS reproduction setup with the probe attached to UR5e manipulator. (b) and (c) Photos of the different phantoms used in the experiments, please refer to the main text for more details.

## V. Experimental Validation

This section outlines the experiments conducted to evaluate the proposed method, starting from the description of the setup and the phantoms. We then describe the specific scenarios tailored for DVT applications. Finally, we describe the metrics used to evaluate the proposed method including the applied force and the US image quality, and compare the performance of the RUS system with the human operator. The supplementary video provides complementary information on the experiments performed, including temporal sequences of US images.

### A. Experimental Setup and Phantoms

The experimental setup consisted of two parts: a demonstration recording setup and a reproduction setup. The demonstration recording setup was described in Section IV-C, and was used to capture the US images, the probe poses and the force data from a human expert performing an US scan.

The reproduction setup replicated the US scan using a UR5e robotic manipulator with its integrated force sensor. The Clarius HD3 Linear US probe was mounted on the robot end-effector using a dedicated 3D printed adapter, as visible in Figure 6 (a). The robotic manipulator was controlled using an admittance controller as described in Section IV-B. The controller parameters were set to the following values:

$$M_p = 2.5\mathbb{I}_3 \quad D_p = 500\mathbb{I}_3 \quad K_p = \mathrm{diag}(270, 270, 0)$$

These parameters were selected to ensure stable tracking of the reference force profiles on the phantoms.

The different devices were interfaced with a software framework specifically developed in Python, which allowed the acquisition and processing of data coming from the different sensors, provided the implementation of the KMP approach adopted, and enabled the control of the robotic manipulator. The drawings and 3D models of the designed adapters, together with the developed software, are publicly available via the following link: https://github.com/lbusellato/CoLeCT.

Two types of phantoms were used in this experiment to simulate the human body with subsurface vessels. The first phantom (Figure 6 (b)) was a box-shaped plastinol phantom with dimensions of $20 \times 18 \times 7.5$ cm. It contained three cylindrical vessels with 10 mm diameter, parallel to the wider surface at depths of 2 cm, 3 cm, and 4 cm from the top surface. The second phantom (Figure 6 (c)) was a softer silicon phantom (TruIV, Limbs&Things co., U.K.) which was commonly used for training in US image-guided vascular procedures. The phantom contained both superficial and deeps veins at a range of diameters (4-8 mm) and in different depths (10-26 mm). During the experiment, the vessels of both phantoms were filled with water in order to assure US imaging quality.

In addition to the above synthetic phantoms, to evaluate the proposed system under more complex and realistic conditions, extra experiments were conducted with human subjects involved. These experiments are described in detail in the next section (see Scenario 3).

### B. Experimental Design

The experimental validation focuses on a US scanning task that requires modulation of interaction forces, inspired by requirements similar to those of a DVT examination. A DVT US scan involves manipulating an US probe along the target vessel while varying pressures are applied to evaluate the presence of blood clots [3]. Within the scope of the described task, we considered three experimental scenarios, each presenting an incremental level of complexity where the RUS was required to acquire US images under different conditions including scanning pressures, positions and directions. In each scenario, the expert demonstrator received verbal scanning instructions, but no additional measures were implemented to ensure data consistency, thereby preserving the natural variability of the demonstrations. To maintain realism, the length of the scanning trajectories was, in all cases, kept consistent with the length of the phantoms (200 mm) and with the typical length of the human forearm (from 250 to 300 mm).

*Scenario 1:* The first scenario was designed to explore KMP's capabilities of generalization and adaptation in a US scanning with approximately constant pressing force. In the experiment, the demonstrator conducted a US scan along the top and middle vessels of Phantom B, spanning an average length of $110 \pm 11$ mm, as well as the superficial vessel of Phantom C, spanning an average length $95 \pm 10$ mm. The pressing force was manually kept constant while ensuring correct coupling between the US probe and the phantom to achieve adequate image quality. Subsequently, KMP was used to produce normal force profiles that replicate the scanning on both the rigid phantom (Phantom B) and the soft phantom (Phantom C). In this scenario, we analyzed the capabilities of KMP to generalize between different phantoms having different biomechanical properties and vessel depths.

*Scenario 2:* The second scenario was designed to reproduce a scanning pattern including vessel compression, while considering the incorporation of waypoints into the force predicted by the KMP. During the demonstration phase, the demonstrator manipulated the US probe (Figure 5) to perform a scan on Phantom C. The scanning trajectory measured $123 \pm 7$ mm on average. The scan commenced with gentle, consistent pressure to ensure optimal coupling between the probe and the phantom. In the middle of the scan, the sliding of the US probe was interrupted. The demonstrator then increased the compression force significantly to collapse the vessel. Subsequently, the KMP method was employed to encode the demonstrations. The consequent reproduction involved a force profile that not only mimicked the demonstrator's actions but also generated a profile with an identical shape yet varied the magnitude of the normal force during the vessel compression. This feature could be clinically advantageous for managing DVT progression, which may necessitate lower compression forces over time to reduce the hemodynamic stress during examinations [3].

*Scenario 3:* The third scenario was designed to explore the use of proposed system in most realistic scanning conditions, and evaluate the system's ability to generalize beyond a set of demonstrations with high inter-demonstration variance. We conducted an experiment involving five healthy male volunteers (average age 29 years, BMI reported in Table III). The demonstrator firstly conducted multiple US scans on volunteers' left arms, aligning the scanning direction with the course of the median vein. The task involved scanning from the elbow towards the wrist in a span of $186 \pm 33$ mm. The movement paused in two places, where compression were performed, while for the rest of the scanning, a relatively constant force was maintained to achieve adequate US image quality. Using appropriate support, we stabilized the forearm parallel to the RUS manipulator's mounting surface to obtain a flat scanning area, both during the demonstration and reproduction.

Each subject is uniquely identified with the abbreviation S1 to S5. The third scenario was designed according to recommendations from the University of Southern Denmark robotics section and the Declaration of Helsinki. The conducted experiments were in strict compliance with internal safety protocols, data management standards, and exempt from ethics committee approval. Before the experiments, all subjects received verbal and written information describing the experiments and its goals. Exploring inter-subject conditions, we trained KMPs on demonstrations from the first four subjects (S1 to S4) and tested on the last subject (S5).

For all the scenarios, as schematically represented in Figure 2, the relative position of the probe was used to predict the optimal interaction force, setting its orientation perpendicular to the flat scanning surface for clear vascular visibility, in line with previous works [2], [20]. We also employed as reference trajectory a linear translation at 10 mm/s aligned with the path of the scanned vessel. Since the scanning conditions were changed in different demonstrations, the position of the scanning start and end point was manually defined before the beginning of each reproduction. Thanks to the application of US gel, the friction on the surface can be considered negligible. By this means, the force was controlled only in the direction normal to the scanning surface, and torques control was not required. Note that this simplification was introduced exclusively to facilitate the presentation of the experimental results, and without loss of generality of the proposed method.

We acquired ten demonstrations for each phantom used in the first two scenarios, while, in Scenario 3, six demonstrations for each healthy volunteer were acquired to reduce experiment times. The KMPs were trained separately for each scenario, using half of the demonstrations (i.e., five data for each phantom, and three data from each human subject), while the rest were used for validation. To assess the efficacy of the conducted experiments, we compared the outcomes from the KMP-driven RUS system with the demonstrations performed by human experts across the various scenarios while training data were excluded to avoid bias. Additionally, the first experiment scenario included control schemes that maintain a constant force, while the second scenario employed a step profile for compression maneuvers. These control schemes could be commonly referenced in prior research [5].

### C. Evaluation Metrics

Our evaluation encompassed both the analysis of interaction forces and the assessment of image quality. Specifically, we analyzed the force profiles normal to the surface over time, summarizing their statistical distribution through mean, standard deviation, maximum, and minimum values.

For the evaluation of image quality, we used an approach similar to [29], [31]. In particular, Peak Signal-to-Noise Ratio (PSNR) and Zero-mean Normalized Cross-Correlation (ZNCC) were used for results comparison. Specifically, we used PSNR to evaluate the image quality in terms of image intensities, where higher values indicate better image quality. In addition, the ZNCC value compared the similarity between a target image and a reference image, where values close to one indicated higher similarity. For both metrics, we considered all the validation acquisitions based on the experts' demonstrations. To evaluate image characteristics over the entire scan, metrics are calculated by comparing all frames in the acquisitions, considering the frames closest in time with respect to the beginning of the scan. Subsequently, we report the average values calculated on all acquisitions.

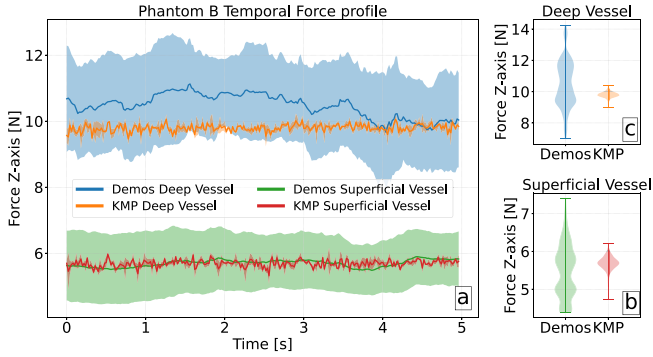

Fig. 7. Temporal profiles (a) and distributions (b,c) of interaction forces obtained in the experiments without compression on both vessels of Phantom B.

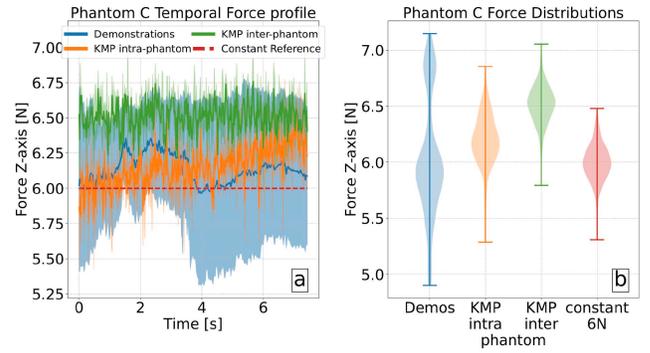

Fig. 8. Temporal profiles (a) and distributions (b) of interaction forces obtained in the experiments without compression on Phantom C.

TABLE I
EXPERIMENTAL DATA (MEAN ± STD) FROM PHANTOM B, CONSIDERING RMSE ERROR FOR FORCES, AND PSNR AND ZNCC IMAGE METRICS

| Phantom Vessel | RMSE ↓ | PSNR ↑ | ZNCC ↑ |
| --- | --- | --- | --- |
| Deep Vessel | 1.49 ± 0.62 | 22.9 ± 0.29 | 0.26 ± 0.03 |
| Superficial Vessel | 0.64 ± 0.28 | 21.12 ± 0.15 | 0.63 ± 0.01 |

## VI. EXPERIMENTAL RESULTS

### A. Scenario 1: Phantom Experiments Without Compression

Firstly, we present the results on the scenario scanning Phantom B with constant pressure, involving demonstrations on both superficial and deep vessels to train the proposed KMP. The temporal force profiles are shown in Figure 7 (a). Our method replicates the force profiles for both vessels, achieving high fidelity for the superficial vessel.

Figures 7 (b,c) display considerable variation in force distribution across all demonstrations for both vessels. In particular, for the deeper vessels, Figure 7 (c), augmented scanning forces exceeding 7 N are observed and peaking at over 14 N, confirming the greater complexity in scanning the deeper vessel compared to the superficial one. Despite this pronounced variability, the KMP obtains a more uniform force profile for both vessels, resulting in an average force of 5.7 N and 9.8 N for the superficial and deep vessels, respectively.

In Table I, we report Root Mean Square Error (RMSE) values for force errors, along with image quality metrics PSNR and ZNCC. The RMSE between KMP force profiles and validation demonstrations is 0.6 ± 0.3 N for the superficial vessel and 1.5 ± 0.6 N for the deeper vessel. In both vessels, the PSNR values exceed 21 dB, indicating an effective coupling between the probe and Phantom B, as the values obtained are comparable to manual acquisitions, which achieve 23.14 ± 0.6. However, low ZNCC values indicate that the images do not closely resemble manual acquisitions due to Phantom B's suboptimal acoustic characteristics. Specifically, Phantom B exhibits high acoustic impedance and a limited presence of backscatters, resulting in weak contrast and low intensity in the US images.

Transitioning to Phantom C, it has reduced stiffness and the vessel has an intermediate depth compared to those in Phantom B. Figure 8 (a) presents the force profile generated by the KMP trained only from demonstration on Phantom B (labeled as the "KMP inter-phantom") and those derived only from demonstrations on Phantom C, referred to as the "KMP intra-phantom". While the KMP intra-phantom yields a force profile that more accurately mirrors the demonstrations, the KMP inter-phantom achieves a competent, albeit marginally elevated, force value. Interaction forces hover around 6 N, serving as a reference. The average RMSE between KMPs and demonstrations is 0.5 ± 0.21 N for "intra-phantom" KMPs and 0.62 ± 0.26 N for "inter-phantom" KMPs. These values are inferior to the constant force reference of 6 N, which stands at 0.72 ± 0.39 N, 30% and 10% larger errors over the aforementioned conditions. Violin plots in Figure 8 (b) highlight irregular force distribution of the demonstrations. While both KMP variants perform well, "intra-phantom" KMPs replicate the demonstrations more faithfully, capturing a broader variability spectrum compared to the KMPs trained on stiffer Phantom B.

In Table II, both quality and similarity indices exceed the threshold of 20 dB for PSNR and 0.8 for ZNCC. The KMP intra-phantom outperforms KMP inter-phantom and the constant force of 6 N. Notably, a constant force reference of 3 N yields the most favorable image quality metrics across all tested approaches. Manual acquisitions achieve an average PSNR of 23.54 ± 0.6 dB and a ZNCC of 0.89 ± 0.01, corresponding to optimal quality and high similarity. The proposed method replicates these values, with intra-phantom KMPs exhibiting negligible discrepancies (errors below 7% for PSNR and 5% for ZNCC). Finding the ideal constant force reference is complex, as metric values tend to diminish incrementally with each 3 N augmentation in the applied force reference.

### B. Scenario 2: Phantom Experiments With Compression

Thus far, our discussion has centered on vessel scanning without addressing the compression task, pivotal for DVT

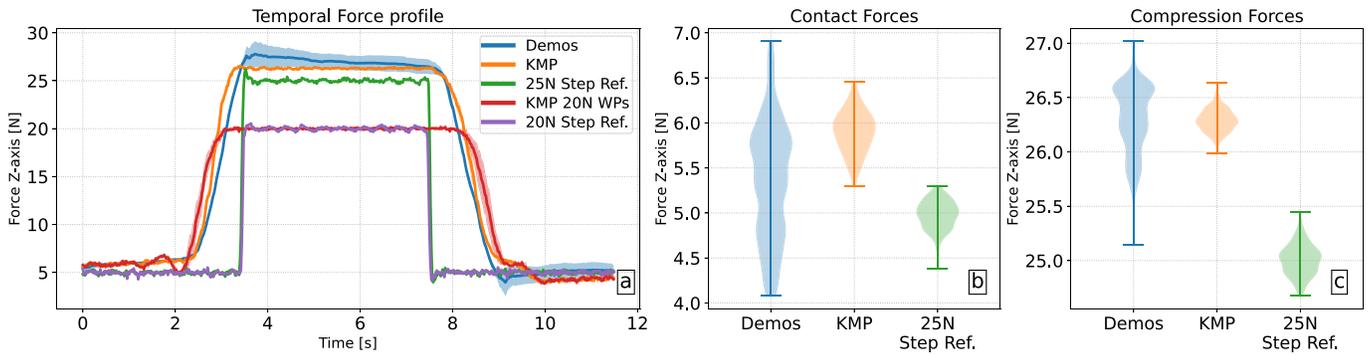

Fig. 9. Temporal profiles (a) and distributions of contact (b) and compression (c) forces obtained in the experiments on anatomical model B.

TABLE II
EXPERIMENTAL DATA (MEAN ± STD) FROM PHANTOM C, CONSIDERING RMSE ERROR FOR FORCES, AND PSNR AND ZNCC IMAGE METRICS

|  | Standard Scan | | |
|---|---|---|---|
|  | **RMSE ↓** | **PSNR ↑** | **ZNCC ↑** |
| KMP inter-phantom | **0.5 ± 0.21** | 20.2 ± 0.18 | 0.8 ± 0.01 |
| KMP intra-phantom | 0.62 ± 0.26 | 22.1 ± 0.29 | 0.85 ± 0.01 |
| 3 N | 3.58 ± 0.52 | **23.02 ± 0.34** | **0.87 ± 0.01** |
| 6 N | 0.72 ± 0.39 | 21.22 ± 0.48 | 0.82 ± 0.01 |
| 9 N | 2.45 ± 0.51 | 20.65 ± 0.18 | 0.79 ± 0.01 |
| 12 N | 5.44 ± 0.52 | 20.24 ± 0.17 | 0.78 ± 0.01 |
|  | Compression Scan | | |
| KMP | **3.93 ± 0.25** | 20.03 ± 0.34 | **0.8 ± 0.01** |
| KMP waypoints | 5.25 ± 0.34 | 19.94 ± 0.23 | 0.79 ± 0.01 |
| 10 N Step | 10.8 ± 0.41 | 19.58 ± 0.29 | 0.76 ± 0.01 |
| 15 N Step | 8.12 ± 0.39 | 19.74 ± 0.28 | 0.76 ± 0.01 |
| 20 N Step | 6.36 ± 0.32 | 19.92 ± 0.25 | 0.77 ± 0.01 |
| 25 N Step | 5.1 ± 0.16 | 19.67 ± 0.33 | 0.75 ± 0.01 |

diagnosis. Compression tests were not feasible on Phantom B due to suboptimal echogenicity and pronounced stiffness. Forces required for the compression during US scanning would exceed 30 N, presenting challenges both in demonstration recording, and reproductions. Phantom C was used for initial compression tests, followed by trials with healthy volunteers.

Figure 9(a) illustrates force profiles achieved on Phantom C, where the demonstration profiles (blue), the reference step profiles with baseline force of 5 N and compression forces of 20 N or 25 N are shown respectively by purple and green curves. KMP profiles (orange) and those with imposed constraints capping compression force at 20 N (red) can accurately emulate the validation demonstrations.

In this compression experiment, we distinguished between the basic contact forces and the compression forces. The force distributions for the initial and terminal phases of the scan, as well as the central compression phase, are depicted in Figures 9(b) and 9(c), respectively. Despite the initial and terminal forces exhibiting significant variation, KMP consistently obtains interaction forces around 5 N, which proved adequate for interacting with Phantom C. KMPs deduce contact forces comparable with those obtained on the same phantom without compression (see results from Scenario 1). In Figure 9(c), KMPs effectively replicate compression force, yielding a concentrated distribution with an average slightly below 26.5 N, despite demonstrations displaying a broader and non-unimodal distribution, peaking around 27 N and barely surpassing 25 N at the minimum. The step 25 N profile does not present comparable force values, even if the force distribution closely resembles the one of KMPs.

The image quality metrics for the compression task (Table II) exhibit a reduction compared to the non-compression task. Manual acquisitions show an average PSNR of 20.5 ± 0.74 dB and a ZNCC of 0.81 ± 0.03. In this compression task, the proposed KMPs (first row) effectively emulate image quality and similarity from manual US scans, even under a maximum force constraint (second row). The KMPs without waypoints surpass the approach based on a 25 N force step, achieving a 7% improvement in PSNR.

The consistency of the experimental results described so far illustrates the ability of the proposed KMP method to maintain a high image quality and force profiles similar to manual demonstrations across variations in anatomical phantoms and task complexities. In addition, the introduction of waypoints allows the adaptation of the force profile to different scanning conditions, a useful element for enabling patient-specific scans. The RUS system reproductions based on KMPs are executed with different scan positions and directions relative to the acquisitions, yet these aspects do not influence the proposed method.

### C. Scenario 3: Healthy Subjects Experiments

Figure 10(a) shows the temporal profiles of normal force during forearm scanning including the demonstrations on five subjects and the KMP executions. Variability exists within each subject and across different subjects, with temporal shifts related to compressions and noticeable differences in forces. Contacting forces during scanning without compression exhibit greater consistency. Notably, the initial force exceeds 5 N, while force between compressions averages less than 5 N, consistent with observations from a single compression on Phantom C. KMP profiles show a clean response during the second compression, while the first compression starts with an initial force of approximately 10 N, peaking at nearly 20 N before stabilizing around 15 N. Considering the higher standard deviation of the demonstrations compared to the phantom experiments, this trend could be linked to the method's attempt to extrapolate the optimal force values for compression. Additionally, we observed that the volunteer exhibited significant nervousness and forearm muscle tension at the beginning of the scan, likely due to a lack of confidence with the RUS system. As the scan progressed, the subject tended to relax, which may partly explain the differences in characteristics between the first and second compressions.

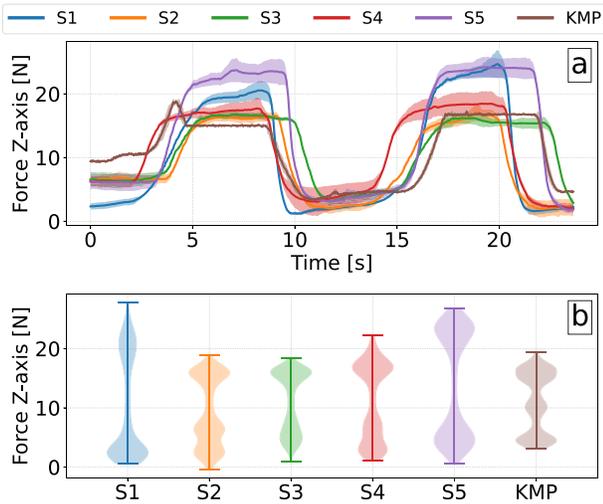

Fig. 10. Contact forces temporal profiles (a) and distribution (b) during forearm scans of healthy volunteers.

TABLE III
KMP RMSE ERRORS AND IMAGE METRICS FOR HEALTHY SUBJECTS EXPERIMENTS

| Ref. Subject | RMSE ↓ | PSNR ↑ | ZNCC ↑ | BMI |
|---|---|---|---|---|
| S1 | 4.95 ± 0.23 | 12.59 ± 0.24 | 0.65 ± 0.03 | 24.5 |
| S2 | 4.87 ± 0.31 | 12.66 ± 0.25 | 0.67 ± 0.02 | 25.6 |
| S3 | 3.55 ± 0.17 | 13.13 ± 0.16 | 0.68 ± 0.02 | 27.8 |
| S4 | 4.7 ± 0.66 | 12.78 ± 0.22 | 0.67 ± 0.01 | 24.2 |
| S5 | 5.56 ± 0.93 | 12.66 ± 0.17 | 0.67 ± 0.02 | 20 |

Violin plots in Figure 10 (b) confirm high variability in interaction force distribution. Despite this, all subject demonstrations exhibit a bimodal distribution corresponding to forces required for scanning without compression (bottom part) and compression phases (top part). Examination of KMP profiles reveals an additional central component absent in demonstrations, with an average value of approximately 10 N corresponding to the initial interaction forces.

In Table III, the KMPs consistently achieve an error below 5 N for subjects S1 to S4. However, subject S5 experiences an error exceeding 5.5 N due to reduced fat mass (BMI index 20). Subject S3 achieves the lowest errors (below 4 N), benefiting from greater adipose tissue facilitating probe-tissue coupling (BMI close to 28).

Image quality metrics in Table III remain consistently low across all subjects. PSNR exceeds 13 dB only for subject S3, while KMP reproduction on subject S5 yields a PSNR of 12.66 which indicates suboptimal image quality. Reduced PSNR values result from the complexity of precise probe-forearm tissue coupling, leading to diminished intensities in US images unrelated to anatomy. Manual acquisitions of the same subject yield an average PSNR of 14.53 ± 0.79 dB, higher than KMP results but lower than Phantom C with easier probe-tissue coupling. ZNCC values consistently remain below 0.7 across all subjects, including subject S5 (ZNCC = 0.67). Low ZNCC reflects significant tissue deformations during scanning due to probe interaction, minor subject's movements, posture changes, and muscle tension. Average ZNCC value for individual subject acquisitions is 0.73 ± 0.03, higher than KMP values but lower than results from the anatomical phantoms.

## VII. DISCUSSION

In this work, we introduced a compact and integrated acquisition setup for recording US scan demonstrations. This approach, tested on anatomical model and healthy subjects considering probe compression during the scan, is ready for broad clinical applications, enabling consistent demonstrations acquisition thanks to shared designs and software. This setup minimally alters the ergonomics of the probe and offers more intuitive demonstration conditions compared to kinesthetic or teleoperated approaches. We plan to validate the proposed design more thoroughly, focusing on ergonomics and ease of use, and compare it with other demonstration approaches in a dedicated future user study.

The proposed acquisition setup, while effective, can be further optimized to handle stress during compression, particularly in design and sensor selection. Most of the available studies consider constant interaction forces required for compression-free scans, limiting the interaction forces to within 20-25 N, with typical forces ranging between 5-10 N [2], [20]. Therefore, we did not initially expect compression forces over 30 N, which can lead to the sleeve deformation and saturate the force sensor range. From the public repository, a refined design can be found featuring a force sensor with a larger measurement range and a mechanical design that increases tolerances while minimizes deformation and clearance, ensuring precise data capture under all conditions.

The results achieved by our proposed method align with or surpass the state-of-the-art techniques in RUS scanning of vascular structures without compression. While direct comparisons may not be entirely fair due to variations in anatomical models and experimental conditions, our findings support that our system's performance meets the operative standards established in prior studies [5], [25], [29], [43]. For example, the force profiles and image quality reported by Goel et al. are similar to our findings, showcasing overlapped ZNCC values and average normal forces between 7.5 and 8.5 N with a standard deviation approximately 2 N [29].

In contrast, IRL approach introduced in [27] reports force errors around 2 N, which is higher than those recorded using our KMP method, even if they consider more intricate anatomical models. In the same work, image quality comparisons are based on the SSIM index, where a forearm scan yields a score of 0.654, closely matching our result of 0.629 calculated on our real subject data. According to [43], SSIM values higher than 0.55 correspond to an adequate quality of US images, confirming the clinical adequacy of the proposed method.

Addressing vessel compression for DVT diagnosis, only a handful of RUS systems take this factor into account. Our experimental validation draws significant inspiration from [5], where the reported force tracking errors without compression are less than 0.6 N, comparable to our KMP-based outcomes. Similarly to their approach, our method requires manual tuning of the control parameters. This could be extended to use a Linear Quadratic Regulator (LQR) to generate the parameters based on the demonstration uncertainty in future work, as has previously been applied in industrial manipulation tasks [44]. An interesting contribution of [5] is their use of a Bounded Barrier-Lyapunov function to ensure convergence of the force error dynamics. Likewise, in our own recent work, we safely bound the interaction forces in an US scanning procedure using a Control Barrier Function [45]. Both approaches can be directly implemented into the existing control scheme. Additionally, direct feedback control on the image quality metrics can be implemented.

The experimental results demonstrate the intrinsic capabilities of the KMP to adapt to different starting and ending points, and to modulate the trajectory in response to external constraints by introducing via-points. These features are fundamental elements to be used in realistic conditions and represent an advancement over previous methods. For instance, it is very complex to implement these features with previous GMM/GMR approaches. Moreover, unlike DMP, KMP can learn efficiently from multiple demonstrations and it requires fewer additional parameters, facilitating its adoption in diverse application contexts. We believe that these characteristics make the KMP an IL approach that allows expert demonstrations to be effectively exploited to increase the autonomy of RUS systems used in the vascular field and beyond.

The best results are achieved when the proposed method replicates the scanning pattern on the same anatomical model. Clinically, this feature can automate single-patient monitoring programs with RUS, evaluating the progression of vascular diseases like DVT and atherosclerosis, while limiting human intervention to the initial demonstrations. Performance is lower when demonstrations and scans are obtained from different models or subjects. This scenario could allow the human operator to scan a limited number of patients and, based on these demonstrations, the RUS system can scan patients not previously acquired. To support this scenario, KMP features such as the introduction of via points or the modification of start and end points could be useful to improve the system adaptation. However, under these conditions it will be necessary to extend the proposed method to consider a larger set of input data.

In fact, the current method presents a degradation of image quality as the complexity of the experimental tasks increases. This aspect may be due to the simplified US scanning conditions considered in this work, where a flat scanning surface is imaged with a fixed perpendicular orientation of the probe. Improvements can be made by adjusting the reference trajectory used in the reproductions. We are considering the adoption of a more sophisticated trajectory planning strategy, similar to the methodology described in [26], [27], or employing a virtual fixture framework as delineated in [5]. Such approaches not only refine the scanning trajectory but also optimize the probe's orientation.

The integration of US images as input to KMPs is probably one of the elements that can most contribute to improving image quality in realistic clinical conditions, and we are actively working on this extension which will be the focus of our research in the near future. Incorporating US images into KMPs may necessitate feature extraction to reduce input dimensionality, a technique previously implemented in BO [32] and in IL [24]. It is important to note that, while these approaches critically rely on dimensionality reduction, KMPs are inherently capable of processing high-dimensional inputs. Nonetheless, feature extraction could facilitate the handling of a larger volume of demonstrations, enabling more efficient and faster training while improving generalization. Furthermore, the integration of deep learning techniques that allow the automatic estimation of US image quality can support exam standardization and more effective clinical diagnosis [46].

The image evaluation metrics currently employed also warrant refinement to integrate semantic information. This enhancement would build upon the methodologies proposed for manual DVT scanning [35], which consider segmentation data, such as the correlation between the applied force and the vessel section visible in US images. While these metrics are straightforward to compute on anatomical phantoms, their clinical application presents challenges, often necessitating time-consuming manual annotations by skilled clinicians. To mitigate this, we are exploring the use of color Doppler data to streamline the generation of segmentation maps [47].

Tests conducted on human volunteers focused on the upper limb, consistent with previous RUS studies [26], [27], [48] also considering DVT [5]. However, this condition is not representative of the majority of DVT disorders, which occur in the lower extremities. Scanning the lower limb introduces multiple complexities, which we are addressing in redesigning the RUS setup to meet the requirements imposed by clinical protocols of US DVT scans. This will require integrating the aspects already described, such as the introduction of US images as input to the KMPs, their training on a more representative and variable set of demonstrations, the automatic generation of more complex scan trajectories that consider the patient's movements and the optimization of the controller to better adapt to different patients while ensuring the necessary safety requirements. In addition to these, we also plan to optimize the patient's positioning relative to the RUS system and the redesign of its end-effector. These enhancements will enable the expansion of our preliminary study, conducted with healthy volunteers, to encompass a patient demographic more indicative of those affected by DVT, including a balanced representation of genders and a broader age range.

## VIII. Conclusion

This work presents an imitation learning-based approach to learn a RUS controller capable of managing probe compression during scanning, as required during the DVT US examination. To improve demonstration ergonomics, we propose an innovative recording device that seamlessly integrates with conventional US probe, adding force-torque and position tracking sensors. During the RUS reproduction, we leverage

Kernelized Movement Primitives to encode the relationship between the scan location and force. Through this, the observed demonstrations can be generalized and adapted to a different application scenario.

Our approach is evaluated extensively on both synthetic phantoms and healthy subjects. Our results show that the KMP method is able to capture the force distribution accurately, resulting in lower RMSE with expert trajectories than constant force baselines, while achieving comparable image metrics (better in the case of compression scans). On live subjects, we have shown that our approach is able to keep the force RMSE under 5 N for four out of five subjects, although image metrics degrade due to variations between the demonstrations and the reproduction conditions.

The findings from this research suggest that our proposed method holds significant potential for the automation of DVT ultrasound examinations by leveraging demonstration-based learning. Nonetheless, for the method to be viable in clinical settings and to ensure consistent image reproducibility, future enhancements should consider more realistic scanning conditions, including probe orientation control and direct integration of US images.

Open Access funding provided by 'Università degli Studi di Verona' within the CRUI CARE Agreement